%% file: main.tex
\pdfoutput=1
\documentclass[letterpaper]{article}
\input{preamble.tex}

\title{T-Lens: Explainable Tool-Call Withholding Beyond Uncertainty\vspace{0.15in}}
\author{
Carlo~Malagnino\textsuperscript{\rm 1},
Xiaoyu~Liang\textsuperscript{\rm 1},
Aliai~Eusebi\textsuperscript{\rm 1},
Virginia~Ceccatelli\textsuperscript{\rm 1},
Tobias~Braun\textsuperscript{\rm 2,1},
Nils~Loose\textsuperscript{\rm 3,1},
Suman~Jana\textsuperscript{\rm 4},
Lorenzo~Cavallaro\textsuperscript{\rm 1,5}
}
\affiliations{\vspace{0.12in}
\textsuperscript{\rm 1}University College London,
\textsuperscript{\rm 2}Technische Universit\"at Darmstadt,
\textsuperscript{\rm 3}Universit\"at zu L\"ubeck\\
\textsuperscript{\rm 4}Columbia University,
\textsuperscript{\rm 5}Mohamed bin Zayed University of Artificial Intelligence
}
\begin{document}
\maketitle

\begin{abstract}
Large language models (LLMs) increasingly act through tools, enabling them to retrieve information and interact with external systems. Reliable tool use requires LLMs to recognize when none of the available tools can serve a request, yet models often make calls even in such situations. A natural safeguard is to withhold calls based on uncertainty estimates. However, on the Berkeley Function-Calling Leaderboard (BFCL), a widely used tool-calling benchmark, uncertainty alone is insufficient to prevent many inappropriate calls. We show that a model's hidden state carries a complementary signal: whether the available tools apply to the request. We introduce \mbox{T-Lens}, a label- and training-free method that withholds inappropriate tool calls before they run. It reads two signals from the model’s hidden state: uncertainty, using U-Space directions, and applicability, using concepts such as necessity and compatibility recovered from the model’s vocabulary. It combines these with output entropy to decide whether to withhold the call. 
On BFCL, T-Lens stops 52\% of inappropriate calls at a 5\% false positive rate, compared with at most 37\% for label-free baselines at the same rate. The advantage carries over to RelyToolBench and MetaTool: using the same directions, T-Lens prevents 55\% and 17\% of inappropriate calls, respectively, compared with at most 23\% and 13\% for baselines requiring no extra model calls.
\end{abstract}

\section{Introduction}
\begin{figure}[!t]
\centering
\includegraphics[width=\columnwidth]{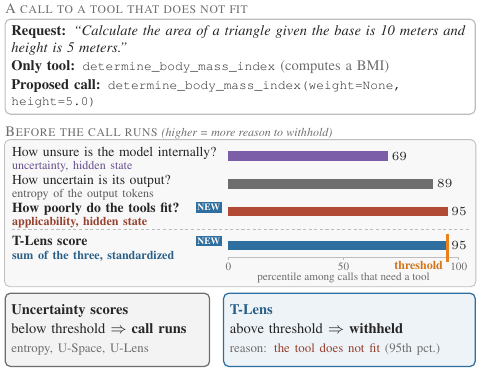}
\caption{\mbox{T-Lens} withholds a call that uncertainty scores let through, and says why. Gemma-3-4B calls a body-mass-index tool to compute a triangle's area. Its uncertainty stays below threshold, but poor tool fit lifts the \mbox{T-Lens} score over it. Bars are percentiles among calls that need a tool. Figure~\ref{fig:pipeline} shows how the signals are computed.}
\label{fig:method}
\end{figure}
Access to external tools allows large language models (LLMs) to retrieve information and act on other systems \citep{schick2023toolformer}. This makes deciding whether to call a tool consequential. An inappropriate call can leak user data or trigger an unwanted action \citep{ruan2024toolemu}. Figure~\ref{fig:method} illustrates a sample failure: asked for the area of a triangle, with only a body-mass-index calculator available, Gemma-3-4B invokes the BMI tool regardless, passing the triangle's height as a person's height. This is not an isolated case: tool-calling benchmarks \citep{patil2025bfcl,ross2025when2call,zhang-etal-2024-toolbehonest} show that models often make such calls.

A common safeguard is to estimate how uncertain the model is before letting it use a tool. Output entropy \citep{malinin2021uncertainty} and G-NLL \citep{aichberger2026rethinking} read the model's token probabilities. \textit{Semantic entropy} \citep{farquhar2024semantic} and \textit{P(True)} \citep{kadavath2022know} query the model again, by sampling several answers or asking it to judge its own, which adds model calls to every decision. The U-Lens \citep{braun2026uspace} reads uncertainty from the hidden state. Trained probes instead predict from hidden states whether a tool is needed or a call is wrong \citep{wu2026tocall,sun2026know,yeats2026calls}, but they need labeled calls and give no reason for a decision. We argue that uncertainty, however it is measured, is not enough. A model may be confident that a tool is necessary while no offered tool fits the request, so a gate must also read whether the tools apply.

The Jacobian lens \citep{gurnee2026workspace} maps an intermediate hidden state to words in the model's vocabulary. Run in reverse, it turns a word into a direction in the hidden state. \citet{braun2026uspace} used this to build U-Space, four directions for ambiguity, incompleteness, conflicting evidence and general uncertainty, each contrasting uncertainty words with their opposites. Directions built this way need no labels, are read from the same forward pass that produces the call, and carry names, so a decision can be explained. We introduce \mbox{T-Lens}, which extends this recipe by contrasting words such as \emph{irrelevant} and \emph{unavailable} with their opposites to obtain a direction for ``this tool does not fit the request''. However, the resulting direction may mix poor tool fit and the model’s uncertainty about the call. We therefore split it, without labels, into an uncertainty component in the U-Space, a novel \emph{applicability} component, which we call the A-Space, and a residual tied to the chosen anchor words, which we discard. 
\mbox{T-Lens} reads the hidden state along both components, adds output entropy, and sums the three standardized scores with equal weights.
Calls on requests that need a tool serve only to calibrate each score’s scale and the threshold, so no classifier is trained and no inappropriate calls are needed.

We evaluate \mbox{T-Lens} on the Berkeley Function-Calling Leaderboard \citep[BFCL;][]{patil2025bfcl}, which offers tools that cannot serve the request and checks whether models refrain. At a matched cost of 5\% necessary calls blocked across 14 BFCL configurations, T-Lens prevents 52\% of inappropriate calls, compared with 35\% for output entropy and 37\% for the U-Lens. It outperforms every label-free baseline at every blocking budget from 1\% to 10\%, and it still catches most of the inappropriate calls that \mbox{U-Lens} blocks. With its directions frozen, \mbox{T-Lens} also transfers to RelyToolBench \citep{xu2025reliability} and MetaTool \citep{huang2024metatool}, where it prevents 55\% and 17\% of inappropriate calls, more than any method that needs no extra model calls (at most 23\% and 13\%). The explanations show whether a withheld call appears to involve a poor tool fit or simply uncertainty about the call. This distinction matters: when applicability does not stand out and the only named reasons are output entropy, general uncertainty, or conflicting evidence, just 51\% of withheld calls are actually inappropriate, compared with 68\% overall.

Our contributions are:
\begin{itemize}
\item \textbf{A label-free gate.}
\mbox{T-Lens} combines uncertainty, applicability, and output entropy to withhold proposed calls before execution, without training a classifier or using examples of inappropriate calls.\footnote{Code and data will be released in a future version.}

\item \textbf{A distinct applicability signal.}
We build the A-Space, orthogonal to the U-Space, from WordNet antonym pairs. Its concepts, such as \emph{necessity} and \emph{compatibility}, recur across models and respond to different kinds of requests than uncertainty does.

\item \textbf{Explanations that reveal reliability.}
For each withheld call, we report which signals stand out. Rejections that name applicability are correct 76\% of the time, while those that name only entropy, general uncertainty, or conflicting evidence are correct 51\% of the time.
\end{itemize}

\begin{figure*}[t]
\centering
\begin{minipage}[t]{0.43\textwidth}
\vspace{0pt}\centering
\includegraphics[width=\linewidth]{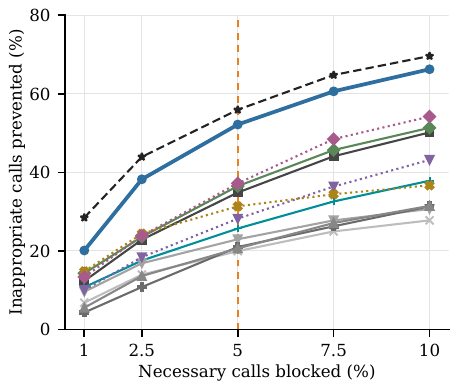}

\vspace{2pt}{\small (a) Matched cost}
\end{minipage}\hfill
\begin{minipage}[t]{0.54\textwidth}
\vspace{0pt}\centering\scriptsize
\input{tables/tab_t2_main.tex}

\vspace{2pt}{\small (b) Calibrated at the 5\% target}
\end{minipage}
\caption{Inappropriate calls prevented on BFCL, as means over 14 configurations. (a)~Matched cost: each method's threshold is set on evaluation data so that it blocks the stated share of necessary calls. The vertical line marks the 5\% budget. (b)~Calibrated evaluation at the 5\% target, in \%, $\pm$ half the width of the 95\% source-group bootstrap interval. Prevented: share of inappropriate calls withheld. Blocked: share of necessary calls (on requests that need a tool) withheld, whether or not the call was right. Correct lost: share withheld among the necessary calls that pass the official tool-and-argument checker, the calls that would have succeeded. The keys in (b) give the line styles in (a). The trained probe, a supervised reference, uses labeled development calls.}
\label{fig:semantic-tradeoff}
\end{figure*}

\section{Background and Related Work}
\paragraph{Tool calling.}
A tool-using model receives the user's request together with a list of tools, each described by a name, a short description and a schema of typed parameters. It either answers in text or emits a call, a tool name with argument values, which a parser decodes and the surrounding system executes. A call is inappropriate when no offered tool serves the request, even if it is well formed. A model can make one because it is unsure, or because it misjudges whether a tool applies. Benchmarks and failure analyses study such calls. BFCL evaluates tool-call correctness and abstention \citep{patil2025bfcl}, and ToolBH whether models recognize tasks that the given tools cannot solve \citep{zhang-etal-2024-toolbehonest}. SABEval traces inappropriate calls to tools whose parameters fit a request they do not serve \citep{liu2026sabeval}, the kind of mismatch that the call in Figure~\ref{fig:method} shows and that our applicability signal targets. Whereas these works measure or explain such failures, \mbox{T-Lens} acts on them at run time: it withholds a call after the model proposes it but before it runs, while controlling how many necessary calls are blocked. Our task is thus a form of selective prediction, closely related to selective question answering, where abstention is known to become harder under domain shift \citep{kamath2020selective}.

\paragraph{Signals from model outputs.}
Uncertainty estimation methods differ mainly in which part of the model they observe: in token probabilities, in distributions over meanings, or in hidden states \citep{liang2026where}. Output entropy measures how unsure the model is about the tokens it generates, while G-NLL scores a single answer by summing the negative log-likelihood of the generated call. \textit{Semantic entropy} instead estimates uncertainty across sampled meanings \citep{farquhar2024semantic}. For function calling, we treat sampled calls as having the same meaning when they parse to the same function and arguments. \citet{ye2026uqfc} find that G-NLL predicts incorrect calls as well as \textit{semantic entropy} does. \textit{P(True)} asks the model to assess its own answer \citep{kadavath2022know}. Both sampling and self-evaluation, however, require additional model calls. Moreover, entropy, G-NLL and semantic entropy quantify how unsure the model is without explaining why, while \textit{P(True)} asks the model directly whether its call is correct. We instead read, from the original generation, two separate signals: both how unsure the model is, and whether any offered tool applies.

\paragraph{Signals from internal representations.}
Trained probes read hidden states to approximate uncertainty \citep{kossen2024probes}, estimate whether a tool is needed \citep{wu2026tocall,sun2026know} or detect tool-call errors \citep{yeats2026calls,ke2026prisms}. However, they require labeled data and training for each model, and they provide no explanation for their decisions. The Jacobian lens \citep{gurnee2026workspace} maps a word to a direction in a model's hidden state, so that projecting a hidden state onto this direction measures how strongly the model represents that word. U-Space \citep{braun2026uspace} uses the lens to construct four orthonormal directions, one for each kind of uncertainty: ambiguity, incompleteness, conflicting evidence and general uncertainty. Each direction contrasts a set of uncertainty words with their opposites (for instance \emph{unclear} versus \emph{clear}, \emph{incomplete} versus \emph{complete}) at a layer roughly two-thirds of the way through the network. The four coordinates of a hidden state thus indicate which kinds of uncertainty are active. U-Space aggregates them into a single score, the cone alignment (a softplus over the four coordinates), and \mbox{U-Lens} multiplies this score by the mean output entropy. Hidden states also reveal decisions beyond uncertainty. \citet{wu2026linear} average the hidden states of example requests for each tool to obtain vectors that predict which tool a model will call, and \citet{kim2026foreword} propose Foreword, which trains predictors of test failures on Jacobian-lens features. Both approaches, however, require examples or labels. The \mbox{T-Lens} applies U-Space's recipe to a mismatch contrast. It retains U-Space’s directions and \mbox{U-Lens}’s use of entropy, and adds an applicability concept that neither contains, thereby combining a hidden-state signal with a token-level one.

\section{T-Lens}
\begin{figure*}[t]
\centering
\includegraphics[width=0.94\textwidth]{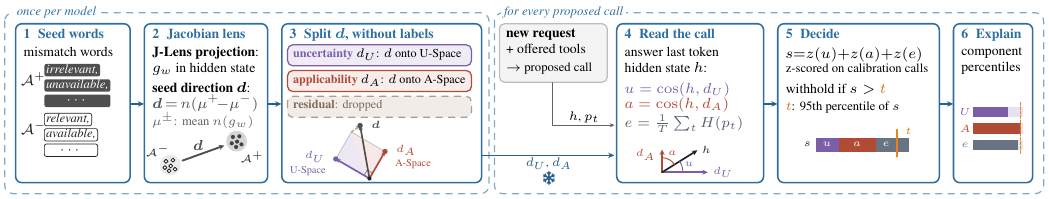}
\caption{How \mbox{T-Lens} works. Once per model and without labels, the Jacobian lens turns mismatch words and their opposites into a seed direction, which is split into an uncertainty component in U-Space, an applicability component in A-Space and a residual that is dropped. For every proposed call, \mbox{T-Lens} reads the hidden state along the two components, adds the mean entropy $e$ of the model's next-token distributions $p_t$ over its $T$ output tokens, and withholds the call when the standardized sum exceeds a threshold calibrated on calls that need a tool.}
\label{fig:pipeline}
\end{figure*}
\paragraph{Seed direction.}
We build one direction for ``this tool does not fit the request'' (Figure~\ref{fig:pipeline}) the way U-Space builds its directions, by contrasting mismatch words with their opposites. Let $J_\ell$ be the mean Jacobian of the Jacobian lens at layer $\ell$, $W_U$ the unembedding and $\gamma$ the final normalization gain. For word $w$, let $g_{w}=J_\ell^\top\operatorname{diag}(\gamma)W_U^\top e_w$ and $n(v)=v/\lVert v\rVert_2$. Six mismatch words $\mathcal A^+=$ \{irrelevant, unrelated, unsupported, unavailable, inappropriate, mismatch\} and their opposites $\mathcal A^-=$ \{relevant, related, supported, available, appropriate, match\} give
\begin{equation}
d=n(\mu^+-\mu^-),\quad \mu^\pm=n\Big(\tfrac{1}{|\mathcal A^\pm|}\textstyle\sum_{w\in\mathcal A^\pm} n(g_{w})\Big),\label{eq:seed}
\end{equation}
at U-Space's layer for each model.

\paragraph{Splitting the direction.}
The seed direction \(d\) contains both uncertainty already represented in the U-Space and a signal of tool applicability, which we identify using pairs of opposite words. We discard the remaining component because it mostly reflects the particular seed words and does not measurably improve detection. Let $Q$ hold an orthonormal basis of U-Space's four directions; the \emph{U-Space component} $d_U$ is the projection of $d$ onto them, one coefficient per category. For the rest, $O=d-d_U$, the candidate concepts are the WordNet 3.0 adjective antonym pairs \citep{miller1995wordnet} whose words are single tokens in every model, excluding the seed pairs. Each pair $(w_k^+,w_k^-)$ gives an axis (429 in all): the difference of its two words' normalized Jacobian-lens directions, with its U-Space component removed,
\begin{equation}
b_k=n\big((I-QQ^\top)\,(n(g_{w_k^+})-n(g_{w_k^-}))\big).\label{eq:axis}
\end{equation}
Since this removal is linear, it gives the same axis as removing the component from each word first. Orthogonal matching pursuit selects the ten axes $b_k$ that best fit $O$. We call their span \emph{A-Space}, by analogy with U-Space. With $Q_A$ an orthonormal basis of A-Space, the \emph{A-Space component} $d_A$ is the orthogonal projection of $O$ onto it:
\begin{equation}
d_U=QQ^\top d,\qquad d_A=Q_AQ_A^\top(d-d_U),\label{eq:split}
\end{equation}
and the residual $d-d_U-d_A$ is removed. These steps use no labels and run once per model.

\paragraph{Decision.}
Each call gets three numbers: how far its hidden state lies along $d_U$, how far along $d_A$, and how unsure the model is about the tokens it generates. The directions $d_U$ and $d_A$ are fixed, computed once per model. Let $h$ be the hidden state at the last token before the answer (after reasoning where present) and $p_1,\dots,p_T$ the model's next-token distributions over its reasoning tokens where present, answer tokens otherwise. Then
\begin{align}
u&=n(d_U)^\top n(h),\qquad a=n(d_A)^\top n(h),\label{eq:reads}\\
e&=\tfrac{1}{T}\textstyle\sum_{t=1}^{T}H(p_t),\label{eq:entropy}\\
s&=z(u)+z(a)+z(e),\label{eq:score}
\end{align}
where $H$ is the entropy and $z$ standardizes each number by its mean and standard deviation on calibration calls to tool-required requests. A call is withheld when $s$ is strictly above a threshold set on the same calls at a 5\% blocking target. The reads $u$ and $a$ (Eq.~\ref{eq:reads}) are the U-Space and A-Space components. Here $u$ reads one fixed blend of U-Space's four categories, with each weighted by how much of it the seed direction contains. This differs from U-Space's own score, introduced above, which combines the four coordinates through the cone alignment and serves as a baseline. Summing (Eq.~\ref{eq:score}) keeps each component's contribution separate, so it can be reported as a reason. It also stops any one component from dominating, as the most variable factor can in a product. The weights are equal because fitting them would need labeled inappropriate calls. This is split-conformal calibration \citep{vovk2005algorithmic,angelopoulos2023gentle}: if calibration and test calls on tool-required requests are exchangeable, a necessary call is blocked with probability at most $(\lfloor\alpha n\rfloor+1)/(n+1)$ for target $\alpha$ and $n$ calibration calls, about 5.2\% here; the bound is only approximate for \mbox{T-Lens} because the same calibration calls also set each score's scale (Appendix~\ref{app:conformal}).

\paragraph{Explanation.}
To explain a rejection, we report how unusual the call is on each of the three components: its percentile among the same calibration calls (Figures~\ref{fig:method} and~\ref{fig:pipeline}). The U-Space component can be broken down further into U-Space's four categories, which tells which kind of uncertainty is involved.

\begin{figure}[t]
\centering
\includegraphics[width=0.92\columnwidth]{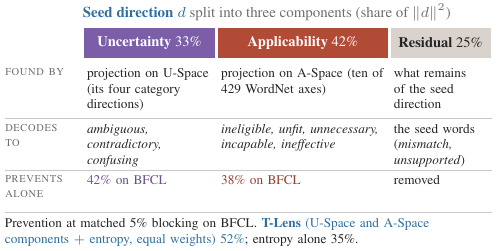}
\caption{The seed direction split into the two concepts \mbox{T-Lens} reads and a removed residual. Shares of $\lVert d\rVert^2$ are means over nine models. Words are the most frequent top Jacobian-lens decodings across models; prevention is at matched 5\% blocking.}
\label{fig:anatomy}
\end{figure}

\begin{figure*}[t]
\centering
\includegraphics[width=0.82\textwidth]{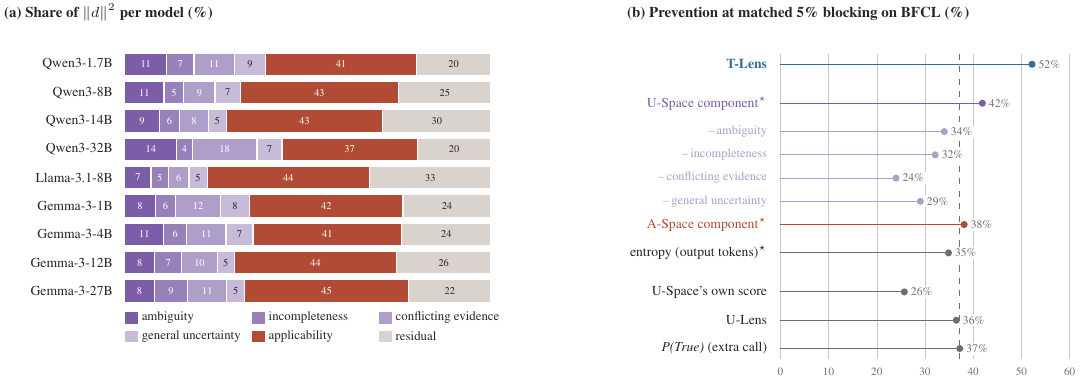}
\caption{(a) Each model's seed direction split into U-Space's four categories, applicability and the residual (shares of $\lVert d\rVert^2$ in \%). (b) Prevention on BFCL at matched 5\% blocking by \mbox{T-Lens} and by each of its components alone: the U-Space component, its four categories, the A-Space component and entropy, with a star on the three that \mbox{T-Lens} combines; U-Space's own score, \mbox{U-Lens} and \textit{P(True)} (extra call) are shown for reference, and the dashed line marks the best of these, \textit{P(True)}.}
\label{fig:uspace-split}
\end{figure*}

\section{Experimental Setup}
\paragraph{Task and costs.}
We gate emitted calls that the benchmark parser can decode; a call is inappropriate when its request is labeled as requiring no call. A gate should prevent as many inappropriate calls as possible while blocking at most a share $\alpha$ of necessary calls (those on requests that do need a tool): prevention is maximized subject to necessary-call blocking $\le\alpha$, with $\alpha=5\%$ fixed in advance as the primary target (Appendix~\ref{app:results}); our conclusions do not depend on this choice, since \mbox{T-Lens} leads every label-free method at every budget from 1 to 10\% (Figure~\ref{fig:semantic-tradeoff}a). With inappropriate calls as positives, prevention is the true-positive rate and blocking the false-positive rate. Our task is whether to call at all: a tool fits when it addresses the request, and we do not judge whether the call's arguments are correct. Arguments still matter for the cost, since withholding a necessary call that would have failed loses little; we therefore also report fully correct-call loss, the share withheld among calls that pass the official tool-and-argument checker.

\paragraph{BFCL.}
We reuse official BFCL v4 single-turn generations and verdicts on 3,035 items: four categories where a tool is needed (simple Python, multiple, live simple, live multiple) and two where no offered tool fits (irrelevance, live irrelevance). The 14 configurations are Qwen3 1.7B/8B/14B/32B in prompt and function-calling modes, Llama-3.1-8B-Instruct in both modes, and Gemma-3 1B/4B/12B/27B in prompt mode; eight have reasoning traces. Within each category, items are split into development (40\%), calibration (20\%) and evaluation (40\%) sets, keeping closely related requests together (near-duplicates, or ones sharing tool documentation; 1,226 source groups). Calibration calls on requests that need a tool set thresholds and the scale of each score, development labels train only a supervised reference probe, and the evaluation set (1,213 items, 12,484 emitted calls) is used only for reporting. For the shift analysis, thresholds calibrated on the expert-curated categories are applied unchanged to the live ones, contributed by users.

\paragraph{External sources.}
To test transfer, we keep each model's directions, layer and sign fixed from BFCL and evaluate six prompt-mode models (Llama-3.1-8B, Qwen3-8B/14B/32B, Gemma-3-12B/27B) on adapted versions of two other benchmarks. In RelyToolBench \citep{xu2025reliability}, each request appears once with tools that can serve it and once with tools that cannot (330 pairs). In MetaTool \citep{huang2024metatool}, each request appears once with a set of plugins that includes a suitable one and once with a set that does not (995 pairs); plugins are offered as functions without arguments, so calling one means choosing it. Each benchmark is split into calibration and evaluation sets that share no sources (Appendix~\ref{app:external}).

\paragraph{Baselines and evaluation.}
We compare with label-free methods of two kinds. From the original generation: output entropy; the entropy and log-probability of the first answer token; the mean answer log-probability and G-NLL \citep{ye2026uqfc}; U-Space's cone-alignment score; and \mbox{U-Lens}, that score times output entropy. With extra model calls: \textit{semantic entropy} with AST clustering of sampled calls \citep{ye2026uqfc}, the share of sampled answers that make no call, and \textit{P(True)}. Entropy uses reasoning tokens where available and answer tokens otherwise; sampling methods use five extra generations. As a supervised reference, a logistic probe on the same hidden state is trained on development labels. We report two settings throughout. At \emph{matched} 5\%, each method's threshold is set on evaluation data so that all block exactly 5\% of necessary calls, which compares ranking; at \emph{calibrated} 5\%, it is set on calibration data alone to target 5\%, as a deployed gate must, and realized blocking can differ. Results are means over configurations; tables give $\pm$ half the width of 95\% intervals from 2,000 paired bootstrap resamples of source groups, computed with the same fixed evaluation code for every method (Appendix~\ref{app:results}).

\section{Results}

\subsection{T-Lens withholds more inappropriate calls}\label{sec:works}
Among label-free methods, \mbox{T-Lens} prevents the most inappropriate calls at every matched budget from 1\% to 10\% (Figure~\ref{fig:semantic-tradeoff}a). At matched 5\%, where every method blocks exactly 5\% of necessary calls, it prevents 52.2\%, against 37.2\% for the best baseline, \textit{P(True)}, 36.5\% for \mbox{U-Lens}, 34.9\% for entropy and 20.6\% for G-NLL; its gains over all four (15.0 to 31.6 points) remain significant after correcting for the thirteen comparisons we test (Bonferroni; Appendix~\ref{app:results}). Without training, it closes about 80\% of the gap between the best label-free baseline and a supervised probe (56.0\%). The advantage is not limited to low budgets: across all thresholds, its AUROC is 0.852, against at most 0.820 for label-free baselines (\textit{P(True)}; Appendix~\ref{app:results}).

At calibrated 5\%, with thresholds set from calibration data alone as in deployment (Figure~\ref{fig:semantic-tradeoff}b), blocking is no longer fixed, so we report both. \mbox{T-Lens} prevents 56.4\% of inappropriate calls while blocking 6.3\% of necessary calls, a balanced accuracy (the mean of prevention and the share of necessary calls let through) of 75.1\%, against 67.9\% for \mbox{U-Lens}, 67.0\% for entropy, 66.8\% for \textit{P(True)}, 58.5\% for G-NLL and 50\% for no gate or a random one. It prevents 12.9, 15.0, 17.4 and 33.6 points more than \mbox{U-Lens}, entropy, \textit{P(True)} and G-NLL respectively, all significant after Bonferroni correction (Appendix~\ref{app:results}), while its blocking differs from theirs (5.6\% to 7.7\%) by no more than sampling error. Some necessary calls would have failed anyway: of those that would have succeeded, \mbox{T-Lens} blocks only 3.6\%, in line with the baselines, so the necessary calls it blocks are disproportionately ones that would have failed.

\subsection{What T-Lens reads: uncertainty and applicability}\label{sec:reads}
The seed direction $d$ splits into three orthogonal components (Figures~\ref{fig:anatomy} and~\ref{fig:uspace-split}): uncertainty (33\% of $\lVert d\rVert^2$ on average), applicability (42\%) and a residual (25\%).

\paragraph{Uncertainty (U-Space component).}
The U-Space component loads on all four of U-Space's categories with the same sign in every model: conflicting evidence takes 10.4\% of $\lVert d\rVert^2$, ambiguity 9.8\%, general uncertainty 6.3\% and incompleteness 6.2\%. Used directly as a score, it prevents 41.9\% at matched 5\% on BFCL, against 25.7\% for U-Space's own cone-alignment score. No single category reaches entropy, but combined they exceed it significantly, by 7.0 points. It correlates moderately with entropy (Spearman 0.35).

\paragraph{Applicability (A-Space component).}
The A-Space component lies in the span of ten WordNet concepts, and the same concepts recur across models: \emph{necessary}/\emph{unnecessary} in all nine, \emph{compatible}/\emph{incompatible} and \emph{present}/\emph{absent} in eight, \emph{eligible}/\emph{ineligible} in six. It decodes to \emph{ineligible}, \emph{unfit}, \emph{unnecessary} and \emph{incapable}: whether a tool can serve the request, rather than whether the model is sure. Alone it prevents 38.1\%, about as much as entropy (34.9\%), while tracking entropy only half as closely as the U-Space component does (Spearman 0.17).

\paragraph{Controlled edits.}
Editing the same requests one way at a time, in four models from three families, shows that the two components respond to different changes (Appendix~\ref{app:decomp}): a tool that cannot serve the request raises applicability in every model, and uncertainty less, while renaming and rewording a fitting tool does neither; an ambiguous argument value raises uncertainty, and applicability much less or, in three of the four, not at all.

\paragraph{Residual.}
The residual decodes to the seed words themselves (\emph{mismatch}, \emph{unsupported}, \emph{unrelated}); dropping it changes prevention by 0.4 points, within noise, so \mbox{T-Lens} removes it (controls in Appendix~\ref{app:results}).

\subsection{Applicability carries over}\label{sec:ablation}
\begin{table}[t]
\centering\footnotesize
\input{tables/tab_ablation.tex}
\caption{Prevention (\%) at matched 5\% blocking for every combination of the three standardized components, summed with equal weights; means over configurations.}
\label{tab:ablation}
\end{table}
Table~\ref{tab:ablation} adds and removes components. On BFCL every component is needed: removing applicability, the U-Space component or entropy costs 4.4, 6.7 and 6.3 points, all significant. Off the benchmark, applicability matters most: without it \mbox{T-Lens} loses 12.0 points on RelyToolBench and 4.6 on MetaTool, more than without the U-Space component, while entropy helps only on BFCL; off it, the two hidden-state components alone do as well. Entropy alone prevents only 15.0\% and 6.9\% there, which suggests that inappropriate calls on these sources are proposed without hesitation in the output tokens: the mismatch is visible in the hidden state but not in the tokens. Applicability is thus the component whose contribution carries over.

\begin{figure}[t]
\centering
\includegraphics[width=0.88\columnwidth]{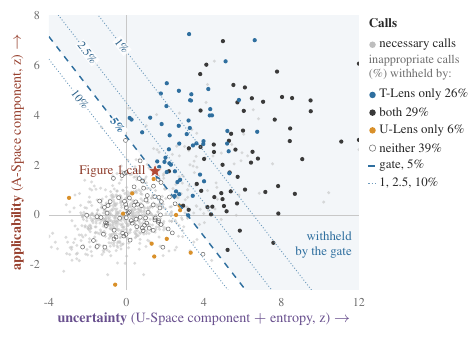}
\caption{Gemma-3-4B evaluation calls in the uncertainty--applicability plane ($z$-scores on calibration calls; star: the Figure~\ref{fig:method} call). \mbox{T-Lens} withholds calls beyond the dashed line, its calibrated 5\% threshold; dotted lines mark 1, 2.5 and 10\%. Colors compare it with \mbox{U-Lens} at its calibrated 5\% threshold; the legend gives shares of inappropriate calls. All configurations: Figure~\ref{fig:plane-all}.}
\label{fig:plane}
\end{figure}

\subsection{Beyond U-Space and U-Lens}\label{sec:plane}
Placing the U-Space component and entropy on one axis (uncertainty) and applicability on the other shows which inappropriate calls each method withholds at its threshold; in this plane \mbox{T-Lens}'s threshold is a straight line (Figure~\ref{fig:plane}). Shares below are pooled over configurations, and medians are per-configuration medians ($z$), averaged.

\paragraph{U-Space.}
With calibrated thresholds, U-Space's score prevents 28.3\% of inappropriate calls and \mbox{T-Lens} 56.4\%, and \mbox{T-Lens} also withholds 90\% of those U-Space's score withholds: adding applicability extends U-Space's catch rather than replacing it.

\paragraph{U-Lens.}
\mbox{U-Lens} multiplies U-Space's score by entropy and ranks calls almost exactly as entropy does (rank correlation 0.98). \mbox{T-Lens} correlates 0.73 with \mbox{U-Lens} and withholds 85\% of the inappropriate calls it withholds. Where the two disagree, the calls have similar uncertainty (median $z$ 3.1 for those only \mbox{U-Lens} withholds, 3.3 for those only \mbox{T-Lens} withholds) but opposite applicability ($-0.3$ against 2.4): the two methods differ only along applicability.

\begin{table}[t]
\centering\footnotesize
\input{tables/tab_explain.tex}
\caption{Rejections at the calibrated 5\% threshold on BFCL (2,001, pooled over 14 configurations): share of rejections that name each reason (a component or U-Space category in its top 5\% among calibration calls to tool-required requests), and share of those that are correct (the call was inappropriate). A gate withholding as many calls per configuration at random would be correct 21\% of the time.}
\label{tab:explain}
\end{table}

\subsection{Explanations flag unreliable rejections}\label{sec:explain}
\mbox{T-Lens} decides with its single score; its components only explain. We name a component, or a U-Space category, as a reason for a rejection (a withheld call) when it alone is in its top 5\% among calibration calls, as extreme as the gate's own threshold (Table~\ref{tab:explain}). For the Figure~\ref{fig:method} call, \mbox{T-Lens} lands at the 95th percentile, just above its threshold, and two reasons are named: applicability (95th) and, within the U-Space component, conflicting evidence (99th), since the only tool computes something else. No uncertainty signal would stop it alone: entropy is high (89th) but below its own threshold, the U-Space component is unremarkable (69th), and U-Space's own score and \mbox{U-Lens} also admit the call.

Reasons also show which rejections to double-check. \mbox{T-Lens}'s rejections are correct 68\% of the time, against 21\% for a gate that withheld as many calls at random. Those that name incompleteness are correct 83\% of the time and those that name applicability 76\%, whereas the fifth that rest only on general uncertainty, conflicting evidence or entropy are correct 51\% of the time, so they most deserve a second look. The score split agrees: rejections in which applicability supplies at least 30\% of the score are correct more often than those in which it supplies under 10\%, by 23.3 points on BFCL and 42.4 on MetaTool (Figure~\ref{fig:triangle}, Appendix~\ref{app:decomp}).

\subsection{Rankings transfer, thresholds less so}\label{sec:transfer}
\begin{table}[t]
\centering\small
\input{tables/tab_transfer_parts.tex}
\caption{Prevention (\%) at matched 5\% blocking with frozen directions; standardization uses each source's calibration calls to tool-required requests. BFCL: 14 configurations; external sources: six models each.}
\label{tab:transfer}
\end{table}
Frozen, \mbox{T-Lens} prevents 55.2\% of inappropriate calls on RelyToolBench and 16.5\% on MetaTool, more than any single-pass baseline (at most 22.8\% and 12.5\%; Table~\ref{tab:transfer}), and ranks calls better than all of them (AUROC 0.86 and 0.76). On RelyToolBench its margin over the best baseline is 32.4 points; the 95\% interval is wide because most models rarely call a tool there when none fits, but its lower end is still 12.8 points. On MetaTool, \mbox{T-Lens} is significantly ahead of every single-pass baseline except U-Space's own score, over which its 3.9-point lead is not significant. By contrast, \textit{P(True)}, which asks the model whether its call is right, prevents twice as much on MetaTool (34.7\%), at the cost of an extra model call for every proposed call. A plausible reason is that MetaTool's task is choosing one plugin among many, a judgment that an explicit question about the proposed call can capture better than a single forward pass.

Thresholds transfer less well than rankings, as is known for selective prediction under distribution shift \citep{kamath2020selective}. Set on curated BFCL requests and applied unchanged to live user queries, the 5\% thresholds overshoot: \mbox{T-Lens} blocks 18.8\% of necessary calls and entropy 15.5\%. \mbox{T-Lens} still prevents 13.8 points more inappropriate calls, but also loses 4.9 points more fully correct calls (Appendix~\ref{app:results}).

\section{Limitations and Future Work}\label{sec:limitations}
\mbox{T-Lens} answers one question on benchmark labels: whether a proposed call should be made at all. The same approach could reach further: reading the hidden state at each argument could flag values that do not fit the request, such as the triangle's height passed as a person's height in Figure~\ref{fig:method}, and in multi-step use the gate would also have to account for earlier calls, their results and the dependencies between them. Beyond BFCL, where it was developed, the evidence comes from two adapted benchmarks; among methods that add no model call, \mbox{T-Lens} prevents the most on all three, although \textit{P(True)} prevents more on MetaTool at the cost of one more call per proposed call. Thresholds, unlike rankings, also need recalibration when the query distribution shifts. Controlled edits show that the two components respond to different changes, though not independently; steering along the applicability direction, as U-Space does for uncertainty, would show whether the model acts on them; if it does, the same direction could turn a withheld call into a better one, toward a fitting tool or a clarifying question, rather than only blocking it. A user study would show whether the explanations help. Finally, the three components are combined with equal weights; optimizing the weighting is left for future work.

\section{Conclusion}
Uncertainty scores miss many calls to tools that do not fit the request. \mbox{T-Lens} adds what they lack: applicability, read from the hidden state without labels. It leads every label-free baseline on BFCL and, frozen, prevents the most among methods without extra model calls on two further benchmarks. Its reasons flag rejections that are often wrong, and applicability is the component that carries over. Whether a model should act depends on how sure it is and on whether the action fits, and its hidden state carries both.

\clearpage
\bibliography{references}

\appendix
\setcounter{secnumdepth}{1}
\section{Conformal Bound}\label{app:conformal}

\textbf{Proposition.} Let $S_1,\dots,S_n$ be gate scores of calls on tool-required calibration requests and $S_{n+1}$ the score of a call on a tool-required test request, with $(S_1,\dots,S_{n+1})$ exchangeable. With calibration scores in decreasing order $S_{(1)}\ge\dots\ge S_{(n)}$, $k=\lfloor \alpha n\rfloor$ and $t=S_{(k+1)}$,
\begin{equation}
P\left(S_{n+1}>t\right)\le \frac{k+1}{n+1},
\end{equation}
with equality when scores are almost surely distinct. By exchangeability the test score's rank among the $n+1$ scores is uniform (ties broken at random), and blocking requires it to rank among the $k+1$ highest. With 209 to 383 calibration calls per configuration, the bound at $\alpha=5\%$ is 5.0 to 5.4\% (mean 5.15\%); for \mbox{T-Lens}, whose standardization is fitted on the same calibration calls, the bound is approximate. Realized blocking on the evaluation split is 6.3\% $\pm$ 2.0 for \mbox{T-Lens}, consistent with the bound, but 7.5\% for entropy and 7.7\% for \mbox{U-Lens}, which fit nothing: calibration and evaluation calls come from different source groups, so exchangeability holds only approximately, and under curated-to-live shift it fails.

\section{Additional Results}\label{app:results}
\begin{table}[!b]
\centering\footnotesize\setlength{\tabcolsep}{3.5pt}
\input{tables/tab_t2_per_model.tex}
\caption{\mbox{T-Lens} per configuration at the calibrated 5\% target: prevented, blocked and correct calls lost (\%), and the difference in prevention from entropy (points).}
\label{tab:semantic-configurations}
\end{table}

\begin{figure*}[t]
\centering
\includegraphics[width=\textwidth]{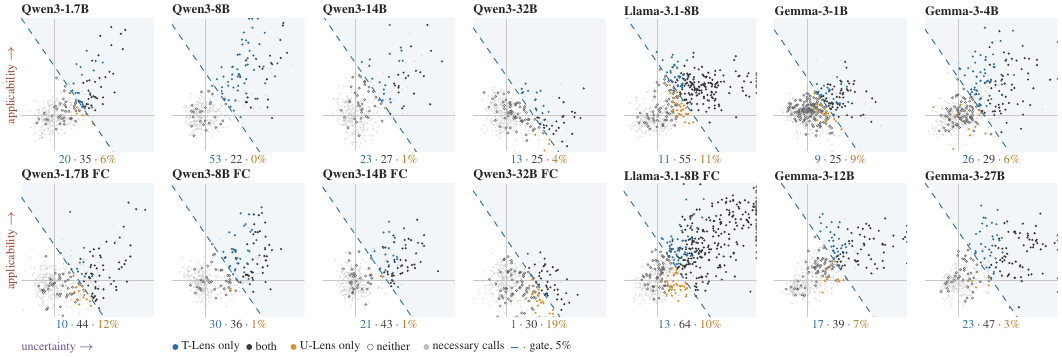}
\caption{The uncertainty--applicability plane (Figure~\ref{fig:plane}) for all 14 configurations, compared with \mbox{U-Lens}, on shared axes (uncertainty $-4$ to 12, applicability $-3$ to 8, in $z$ units). Each panel shows a fixed random subsample of 300 necessary calls and all inappropriate calls. Numbers under each panel: shares of inappropriate calls withheld only by \mbox{T-Lens}, by both, and only by \mbox{U-Lens}.}
\label{fig:plane-all}
\end{figure*}

\paragraph{Per-configuration results and statistics.}
Table~\ref{tab:semantic-configurations} gives per-configuration results (\mbox{T-Lens} prevents more than entropy in eleven of the fourteen configurations) and Table~\ref{tab:semantic-shift} the curated-to-live shift. Intervals use 2,000 paired source-group draws, Bonferroni-adjusted within the matched-cost family and, separately, the calibrated family of twenty-four contrasts. Sixteen BFCL calls come from Qwen3 generations that reached the 4,096-token limit before closing their reasoning, so they have no reasoning tokens to take entropy over; they receive the calibration mean.

\paragraph{Budget.}
The 5\% budget was fixed in advance; on held-out calls, each extra necessary call blocked stops more than one extra inappropriate call only up to about 4\% of blocking, so with equal costs 4 to 5\% is the break-even point.

\paragraph{AUROC.}
Table~\ref{tab:auroc} gives AUROC: \mbox{T-Lens} is highest among methods that use only the original generation on all three sources, and \textit{P(True)}, with an extra model call, is higher on MetaTool. AUROC averages over blocking levels far above the target, which is why we compare methods at matched budgets.

\begin{table}[t]
\centering\footnotesize\setlength{\tabcolsep}{3pt}
\resizebox{\columnwidth}{!}{\input{tables/tab_t2_shift.tex}}
\caption{Curated calibration to live-query evaluation, thresholds frozen at the 5\% target; definitions as in Figure~\ref{fig:semantic-tradeoff}b.}
\label{tab:semantic-shift}
\end{table}

\begin{table}[t]
\centering\footnotesize
\input{tables/tab_auroc.tex}
\caption{AUROC for separating inappropriate from necessary evaluation calls, means over configurations, for label-free methods that use only the original generation and, below the rule, \textit{P(True)}, which needs an extra model call.}
\label{tab:auroc}
\end{table}

\paragraph{Controls.}
We pass 200 random seed sets, drawn from a task-related word pool, through the same pipeline with entropy included: their mean calibrated prevention is 33.6\%, against 56.4\% for \mbox{T-Lens}, and only three of the 200 match or exceed it. Seeds of negated adjectives unrelated to tools (\emph{happy}/\emph{unhappy}) come closer, 53.8\% on average over 50 sets, with five of them at or above \mbox{T-Lens}: the applicability concepts are themselves negated adjectives (\emph{unfit}, \emph{ineligible}), and much of what the gate reads is shared with such generic negative evaluations.

\paragraph{Readout position.}
Reading the two components at the end of the prompt instead of just before the answer, with entropy unchanged, lowers matched prevention by 8.6 $\pm$ 5.4 points on the eight reasoning configurations: the hidden state carries more of the signal once the model has reasoned.

\section{Decomposition Details}\label{app:decomp}
\begin{table}[!b]
\centering\footnotesize
\resizebox{\columnwidth}{!}{\input{tables/tab_anatomy.tex}}
\caption{Prevention on BFCL at matched 5\% blocking of the seed direction, each of its components, entropy and \mbox{T-Lens}, means over 14 configurations, with differences from entropy $\pm$ half the width of their paired 95\% source-group bootstrap intervals (2,000 draws).}
\label{tab:anatomy}
\end{table}

\begin{figure}[t]
\centering
\includegraphics[width=\columnwidth]{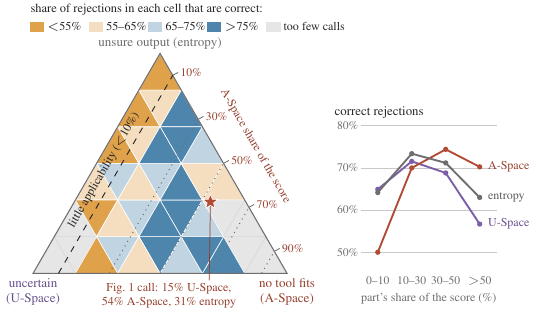}
\caption{Rejections at the calibrated 5\% threshold on BFCL (14 configurations), placed by how each score splits between the U-Space component, the A-Space component and entropy (the corners). Cells show the share of rejections that are correct (gray: fewer than ten); the line chart plots it against each component's share of the score.}
\label{fig:triangle}
\end{figure}

\paragraph{What each component prevents.}
Table~\ref{tab:anatomy} gives each component on its own at matched 5\% on BFCL. The U-Space component prevents 7.0 points more than entropy, and the A-Space component about as much as entropy. No single U-Space category beats entropy, although incompleteness, which correlates least with entropy (Spearman 0.14), costs the most when removed from the U-Space component (41.9\% to 36.8\%). The seed direction prevents 46.6\% alone and 50.1\% with entropy added; \mbox{T-Lens}, which sums the two components and entropy, reaches 52.2\%. Splitting the direction thus adds only 2.1 points on BFCL, 2.4 on RelyToolBench and 2.9 on MetaTool: its main contribution is the explanation.

\paragraph{Shares and decodings.}
The three components are mutually orthogonal; across models they hold 22 to 43\% (U-Space), 37 to 45\% (A-Space) and 20 to 33\% (residual) of $\lVert d\rVert^2$. We decode a component by ranking the Jacobian-lens directions of 12,645 single-token WordNet lemmas by cosine with it; each U-Space axis decodes to its own category's anchors. The A-Space axes come from WordNet's antonym pairs. Besides the concepts in the main text, \emph{affected}/\emph{unaffected} is selected in all nine models but has no clear interpretation; it carries 4 to 9\% of the A-Space component, and excluding it from the candidate axes changes prevention by at most 1.2 points. For Qwen3-32B, the U-Space and A-Space components decode to unrelated words (\emph{uglify}, \emph{semicolon}), although its selected axes (\emph{eligible}/\emph{ineligible}, \emph{necessary}/\emph{unnecessary}) match the other models'.

\paragraph{Seed words versus applicability.}
The seed words name a mismatch directly; applicability is broader: whether a tool is \emph{necessary}, \emph{compatible} or \emph{eligible} for the request. On average 4.6 of the six seed risk words are among the residual's 25 top decodings per model, against 1.7 for applicability and 0.1 for uncertainty. What is specific to the chosen words therefore ends up in the residual, which is removed, while applicability is the part of the seed direction shared with a wider vocabulary.

\begin{figure}[t]
\centering
\includegraphics[width=\columnwidth]{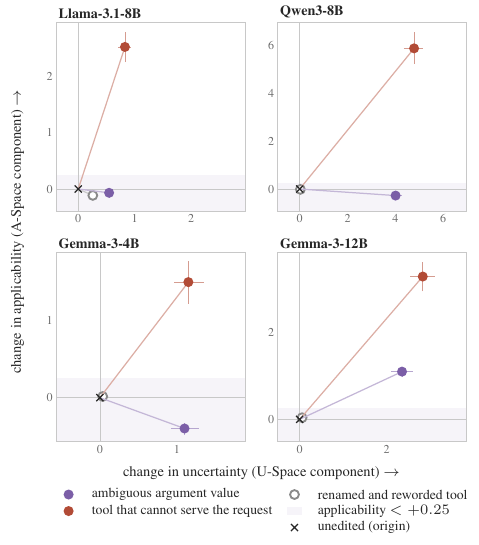}
\caption{Controlled edits, one panel per model: change in the uncertainty and applicability reads from the unedited (or certain) version, when an argument value is made ambiguous, the tool is replaced by one that cannot serve the request, or the correct tool is renamed and reworded. Equal units on both axes within each panel (standard deviations of the reads over the unedited requests); crosses: 95\% intervals over requests.}
\label{fig:edits}
\end{figure}

\paragraph{Controlled edits.}
We edit 192 BFCL requests one way at a time and read the hidden state just before the answer in Llama-3.1-8B, Qwen3-8B (in its non-thinking mode), Gemma-3-4B and Gemma-3-12B, measuring each change in standard deviations of the reads over the unedited requests. To vary applicability with the request fixed, we replace the correct tool's name and description, keeping its parameters, with a function from the same domain that cannot serve the request (for example, a hotel availability check for a booking request); as a control, we rename and reword the correct tool without changing its meaning. To vary uncertainty with the correct tool offered, we make one argument value ambiguous without uncertainty words: ``2 (2 or 4) nights'' against ``2 (2, not 4) nights''. The unfitting tool raises applicability by 1.5 to 5.9 in all four models, and uncertainty by less; the control moves either by at most 0.3 (Figure~\ref{fig:edits}). The ambiguous value raises uncertainty by 0.5 to 4.0. Applicability does not rise in three models (the upper end of its 90\% interval stays below 0.25); in Gemma-3-12B it rises by about half as much as uncertainty, consistently across requests, as if this model also registers that no single call can then be fully correct. In every model, applicability takes a larger share of the change for the unfitting tool than for the ambiguous value. The two components thus respond to different changes, though not independently.

\paragraph{All configurations.}
Figure~\ref{fig:plane-all} repeats Figure~\ref{fig:plane} for every configuration; Figure~\ref{fig:triangle} shows how BFCL rejections split between components.

\section{External Sources}\label{app:external}
Both sources are our adaptations, not the benchmarks' official settings, and each request appears once with tools that can serve it and once with tools that cannot. RelyToolBench, derived from StableToolBench, has 348 such pairs; we drop eighteen, before seeing any results, whose ``no tool fits'' version still offered a relevant tool. The remaining 660 inputs form 191 source groups (57 for calibration, 134 for evaluation), and only the first call a model proposes is scored. Most models rarely call a tool when none fits: of the 203 such evaluation requests, they do so 6 to 11 times, and Llama-3.1-8B 68 times. \mbox{T-Lens}'s margin rests mostly on Llama-3.1-8B, the model with the most such calls (76.5\% against 48.5\% for the best baseline); on Qwen3-32B, whose decodings are uninformative (Appendix~\ref{app:decomp}), it prevents none of its seven. MetaTool contributes 995 pairs (Task~2, Subtasks~1 and~3), each offering ten plugins, split by target plugin into 59 calibration and 140 evaluation groups; we shuffle the plugins so the suitable one is not always listed first. In both, a tool counts as fitting when the benchmark marks it relevant, which does not establish that calling it is necessary or that its arguments are right. Intervals use 2,000 paired source-group draws, Bonferroni-adjusted over twelve contrasts against entropy and unadjusted (95\%) against the other single-pass baselines.

\end{document}

%% file: preamble.tex
\usepackage[preprint]{aaai2027}
\usepackage[hyphens]{url}
\usepackage{graphicx}
\usepackage{natbib}
\usepackage{caption}
\usepackage{booktabs}
\usepackage{amsmath,amssymb}
\usepackage{tabularx,array}
\newcolumntype{L}{>{\raggedright\arraybackslash}X}

%% file: tables/tab_t2_main.tex
\begin{tabular}{@{}lrrr@{}}
\toprule
Method & Prevented $\uparrow$ & Blocked $\downarrow$ & Correct lost $\downarrow$ \\
\midrule
\multicolumn{4}{@{}l}{\textit{Label-free, original generation only}} \\
\raisebox{.25ex}{\includegraphics{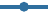}}\,\textbf{T-Lens} & \textbf{56.4 $\pm$ 5.2} & \textbf{6.3 $\pm$ 2.0} & \textbf{3.6 $\pm$ 1.1} \\
\raisebox{.25ex}{\includegraphics{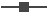}}\,Entropy & 41.4 $\pm$ 4.1 & 7.5 $\pm$ 1.6 & 4.0 $\pm$ 0.9 \\
\raisebox{.25ex}{\includegraphics{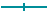}}\,U-Space & 28.3 $\pm$ 3.3 & 5.9 $\pm$ 1.1 & 5.2 $\pm$ 0.9 \\
\raisebox{.25ex}{\includegraphics{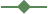}}\,U-Lens & 43.5 $\pm$ 4.4 & 7.7 $\pm$ 1.6 & 4.0 $\pm$ 0.9 \\
\raisebox{.25ex}{\includegraphics{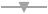}}\,First-token entropy & 25.7 $\pm$ 3.2 & 7.2 $\pm$ 1.5 & 6.5 $\pm$ 1.5 \\
\raisebox{.25ex}{\includegraphics{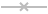}}\,First-token log-prob. & 22.4 $\pm$ 3.0 & 7.3 $\pm$ 1.3 & 6.7 $\pm$ 1.3 \\
\raisebox{.25ex}{\includegraphics{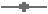}}\,Mean answer log-prob. & 24.8 $\pm$ 3.0 & 6.8 $\pm$ 1.2 & 5.4 $\pm$ 0.9 \\
\raisebox{.25ex}{\includegraphics{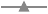}}\,G-NLL & 22.9 $\pm$ 3.4 & 5.8 $\pm$ 0.9 & 3.9 $\pm$ 0.5 \\
\midrule
\multicolumn{4}{@{}l}{\textit{Label-free, extra model calls}} \\
\raisebox{.25ex}{\includegraphics{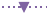}}\,\textit{Semantic entropy} & 20.6 $\pm$ 4.3 & 3.7 $\pm$ 0.9 & 1.3 $\pm$ 0.4 \\
\raisebox{.25ex}{\includegraphics{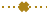}}\,Sampled no-call rate & 26.5 $\pm$ 3.8 & 2.6 $\pm$ 0.7 & 1.4 $\pm$ 0.4 \\
\raisebox{.25ex}{\includegraphics{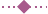}}\,\textit{P(True)} & 39.1 $\pm$ 5.5 & 5.6 $\pm$ 1.3 & 2.9 $\pm$ 0.9 \\
\midrule
\multicolumn{4}{@{}l}{\textit{Supervised reference}} \\
\raisebox{.25ex}{\includegraphics{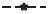}}\,Trained probe & 61.3 $\pm$ 5.8 & 6.7 $\pm$ 1.8 & 4.2 $\pm$ 1.1 \\
\bottomrule
\end{tabular}

%% file: tables/tab_ablation.tex
\setlength{\tabcolsep}{4pt}\begin{tabular}{@{}lrrr@{}}
\toprule
Prevention (\%) & BFCL & RelyToolBench & MetaTool \\
\midrule
U-Space component & 41.9 & 58.0 & 18.8 \\
A-Space component & 38.1 & 50.6 & 18.6 \\
Entropy & 34.9 & 15.0 & 6.9 \\
\midrule
U + A (no entropy) & 45.9 & 58.1 & 18.6 \\
U + entropy (no A) & 47.8 & 43.1 & 11.8 \\
A + entropy (no U) & 45.5 & 48.2 & 13.9 \\
\midrule
\textbf{\mbox{T-Lens}} (all three) & 52.2 & 55.2 & 16.5 \\
\bottomrule
\end{tabular}

%% file: tables/tab_explain.tex
\setlength{\tabcolsep}{4pt}\begin{tabular}{@{}lrr@{}}
\toprule
Rejections that\ldots & Share (\%) & Correct (\%) \\
\midrule
name incompleteness & 42 & 83 \\
name applicability & 58 & 76 \\
name ambiguity & 45 & 73 \\
name entropy & 64 & 70 \\
name conflicting evidence & 45 & 67 \\
name general uncertainty & 39 & 66 \\
\midrule
rest only on general uncertainty, & & \\
\quad conflicting evidence or entropy & 22 & 51 \\
all rejections & 100 & 68 \\
random, same number & & 21 \\
\bottomrule
\end{tabular}

%% file: tables/tab_transfer_parts.tex
\setlength{\tabcolsep}{4pt}\begin{tabular}{@{}lrrr@{}}
\toprule
Prevention (\%) & BFCL & RelyToolBench & MetaTool \\
\midrule
\textbf{\mbox{T-Lens}} & \textbf{52.2} & \textbf{55.2} & \textbf{16.5} \\
U-Space & 25.7 & 15.9 & 12.5 \\
\mbox{U-Lens} & 36.5 & 15.7 & 7.5 \\
Entropy & 34.9 & 15.0 & 6.9 \\
First-token entropy & 23.0 & 22.8 & 11.1 \\
First-token log-prob. & 20.0 & 16.1 & 10.3 \\
Mean answer log-prob. & 21.0 & 19.5 & 7.8 \\
G-NLL & 20.6 & 8.7 & 11.2 \\
\midrule
\textit{P(True)} (extra call) & 37.2 & 19.2 & 34.7 \\
\bottomrule
\end{tabular}

%% file: tables/tab_t2_per_model.tex
\begin{tabular}{@{}lrrrr@{}}
\toprule
Configuration & Prev. & Block. & Lost & $\Delta$ prev. \\
\midrule
Qwen3-1.7B (prompt) & 54.5 & 6.5 & 3.8 & 11.9 \\
Qwen3-1.7B-FC & 54.0 & 7.6 & 4.1 & -3.2 \\
Qwen3-8B (prompt) & 74.5 & 7.0 & 4.4 & 52.9 \\
Qwen3-8B-FC & 66.4 & 5.2 & 2.6 & 33.6 \\
Qwen3-14B (prompt) & 50.0 & 4.5 & 1.4 & 24.4 \\
Qwen3-14B-FC & 64.0 & 7.2 & 3.5 & 29.0 \\
Qwen3-32B (prompt) & 38.7 & 5.7 & 3.3 & 7.5 \\
Qwen3-32B-FC & 31.3 & 5.0 & 3.2 & -13.4 \\
Llama-3.1-8B (prompt) & 66.0 & 4.2 & 1.9 & 1.1 \\
Llama-3.1-8B-FC & 76.3 & 5.7 & 1.9 & 10.8 \\
Gemma-3 1B (prompt) & 33.8 & 5.8 & 1.1 & 0.0 \\
Gemma-3 4B (prompt) & 55.2 & 10.9 & 8.3 & 23.1 \\
Gemma-3 12B (prompt) & 55.9 & 4.6 & 2.9 & 9.9 \\
Gemma-3 27B (prompt) & 69.5 & 8.4 & 7.5 & 22.5 \\
\bottomrule
\end{tabular}

%% file: tables/tab_t2_shift.tex
\begin{tabular}{@{}lrrr@{}}
\toprule
Method & Prevented (\%) $\uparrow$ & Blocked (\%) $\downarrow$ & Correct lost (\%) $\downarrow$ \\
\midrule
\textbf{T-Lens} & \textbf{72.3 $\pm$ 2.8} & \textbf{18.8 $\pm$ 2.3} & \textbf{14.0 $\pm$ 1.8} \\
Entropy & 58.5 $\pm$ 3.4 & 15.5 $\pm$ 1.9 & 9.1 $\pm$ 1.2 \\
U-Space & 36.0 $\pm$ 2.5 & 11.2 $\pm$ 1.2 & 10.1 $\pm$ 1.1 \\
U-Lens & 60.6 $\pm$ 3.8 & 15.6 $\pm$ 2.0 & 9.3 $\pm$ 1.2 \\
G-NLL & 34.5 $\pm$ 3.7 & 10.5 $\pm$ 1.3 & 6.3 $\pm$ 0.8 \\
\midrule
\textit{P(True)} (extra call) & 62.6 $\pm$ 3.9 & 18.7 $\pm$ 2.2 & 13.7 $\pm$ 1.7 \\
\midrule
\multicolumn{4}{@{}l}{\textit{Supervised reference}} \\
Trained probe & 76.0 $\pm$ 3.4 & 20.8 $\pm$ 2.3 & 17.1 $\pm$ 1.8 \\
\bottomrule
\end{tabular}

%% file: tables/tab_auroc.tex
\setlength{\tabcolsep}{4pt}\begin{tabular}{@{}lrrr@{}}
\toprule
AUROC & BFCL & RelyToolBench & MetaTool \\
\midrule
\textbf{\mbox{T-Lens}} & \textbf{0.852} & \textbf{0.862} & \textbf{0.755} \\
U-Space & 0.726 & 0.798 & 0.739 \\
\mbox{U-Lens} & 0.810 & 0.706 & 0.612 \\
Entropy & 0.800 & 0.689 & 0.600 \\
First-token entropy & 0.665 & 0.688 & 0.582 \\
First-token log-prob. & 0.634 & 0.675 & 0.568 \\
Mean answer log-prob. & 0.693 & 0.680 & 0.574 \\
G-NLL & 0.685 & 0.637 & 0.602 \\
\midrule
\textit{P(True)} (extra call) & 0.820 & 0.731 & 0.773 \\
\bottomrule
\end{tabular}

%% file: tables/tab_anatomy.tex
\setlength{\tabcolsep}{4pt}\begin{tabular}{@{}lrr@{}}
\toprule
BFCL, matched 5\% & Prevented (\%) & vs.\ entropy (pp) \\
\midrule
Seed direction & 46.6 & \\
U-Space component & 41.9 & +7.0 $\pm$ 5.3 \\
\quad ambiguity alone & 34.0 & $-$0.9 $\pm$ 5.6 \\
\quad incompleteness alone & 32.1 & $-$2.8 $\pm$ 4.9 \\
\quad conflicting evidence alone & 24.0 & $-$10.9 $\pm$ 4.1 \\
\quad general uncertainty alone & 29.0 & $-$5.8 $\pm$ 5.4 \\
A-Space component & 38.1 & +3.3 $\pm$ 5.9 \\
Residual & 29.1 & \\
Entropy & 34.9 & \\
\textbf{\mbox{T-Lens}} (all three) & \textbf{52.2} & \\
\bottomrule
\end{tabular}